\documentclass[11pt,letterpaper]{article}
\usepackage[letterpaper,vmargin=25mm,hmargin=12.5mm,headheight=40pt]{geometry}
\usepackage{mathptmx}
\usepackage{helvet}                       % Arial substitute
\usepackage{longtable}

\usepackage{amsmath,amssymb}
\usepackage{graphicx}
\usepackage{booktabs}
\usepackage{fancyhdr}
\usepackage{titlesec}
\usepackage{caption}
\usepackage[hidelinks]{hyperref}

\newcommand{\papernum}{104}
\titleformat{\section}{\sffamily\bfseries\fontsize{12}{14}\selectfont}{\thesection.}{0.5em}{}
\titleformat{\subsection}{\sffamily\bfseries\fontsize{11}{13}\selectfont}{\thesubsection}{0.5em}{}
\titlespacing{\section}{0pt}{11pt}{0pt}
\titlespacing{\subsection}{0pt}{11pt}{0pt}

\fancypagestyle{firstpage}{%
  \fancyhf{}
  \fancyhead[L]{\footnotesize
    Proceedings of the 12th International Conference on Combustion Science and Processes (CSP 2027)\\
    London, United Kingdom -- April, 2027\\
    Paper No. CSP \papernum\\
    DOI: 10.11159/csp27.\papernum}
  \fancyfoot[C]{\small CSP \papernum{}-\thepage}
  
}

\newcommand{\figplaceholder}[3]{% {filename}{width}{placeholder text}
  \IfFileExists{#1}{\includegraphics[width=#2]{#1}}%
  {\fbox{\parbox[c][3.2cm][c]{0.85\linewidth}{\centering\small [#3 -- insert \texttt{\detokenize{#1}}]}}}}

\begin{document}

\thispagestyle{firstpage}

\begin{center}
  {\sffamily\bfseries\fontsize{16}{19}\selectfont Physics-Guided Generative Modelling of Oxy-Combustion Flame Images: From Conditional LSGANs to Denoising Diffusion Models}\\[14pt]

  {\bfseries\fontsize{12}{14}\selectfont Imane Er-rabhi$^{1,2}$, Gorkem Oztarlik$^{1}$, Jean Caudal$^{1}$,  Renaud Gaudron$^{1}$}\\[4pt]

  {\small
  $^{1}$Air Liquide, Paris Innovation Campus\\
  1 Chemin de la Porte des Loges, 78350 Les Loges-en-Josas, France\\
  imane.er-rabhi@grenoble-inp.org; gorkem.oztarlik@airliquide.com; jean.caudal@airliquide.com;\\ renaud.gaudron@airliquide.com;\\[3pt]
  $^{2}$Grenoble INP -- Ensimag, Universit\'e Grenoble Alpes, 
  Grenoble, France
  }
\end{center}

\vspace{6pt}

\noindent{\small{\sffamily\bfseries\itshape Abstract -} \textit{Validating industrial oxy-combustion burner designs traditionally requires costly experimental campaigns in which each operating condition is physically reproduced on an instrumented rig and imaged via OH* chemiluminescence, rendering iterative design exploration prohibitively slow. This work develops deep generative surrogate models that map a 15-dimensional vector of continuous operating parameters (thermal power, swirl numbers, momentum ratios, mixture fractions, etc.) directly to a $512\times64$ heat-release flame image, trained on 1,609 experimental samples from the OXYTEC dual-swirl non-premixed oxy-combustion database. Two complementary generative paradigms are investigated, as they offer opposite strengths for surrogate modelling: a custom spatially-anchored conditional least-squares GAN (cLSGAN-L1-CBN), whose single-pass generation enables near-instantaneous inference, and a custom physics-guided denoising diffusion probabilistic model (DDPM), whose iterative sampling favours the smooth, continuous heat-release gradients of fluid flows. Both are validated through systematic ablation on a stratified held-out test set. The optimized cLSGAN reaches RMSE 0.022 and FID 20.68 and generates an image in about 20\,ms on a mobile workstation GPU, though extended training leaves its texture realism unchanged; the DDPM improves steadily over long training horizons, reaching RMSE 0.019 and FID 20.29 at 40,000 epochs, at the cost of roughly thirty seconds per image on the same hardware --- three orders of magnitude slower than the adversarial surrogate. A nearest-neighbour analysis traces the persistent worst-case errors to sparsely sampled regions of the parameter space, and a parameter-sensitivity study confirms smooth, physically consistent responses to perturbed inputs. These results demonstrate that carefully guided generative models can act as reliable visual surrogates for combustion experiments.}}

\vspace{4pt}

\section*{Nomenclature}
\vspace{-2mm}
{\small
\begin{longtable}{@{}ll@{\hspace{6mm}}ll@{}}
\multicolumn{4}{@{}l}{\textit{Operating parameters (network inputs)}}\\[1mm]
$J$ & equivalent momentum flux ratio, $\rho_2^{\mathrm{eq}}u_2^2/(\rho_1u_1^2)$ &
$P$ & total burner thermal power [kW]\\
$\Phi$ & global equivalence ratio &
$\theta_q$ & quarl angle [$^\circ$]\\
$Z_{st}$ & mixture fraction at stoichiometry &
$T_{eq,st}$ & equilibrium temperature at stoichiometry [K]\\
$X_{st}$ & mole fraction of the inner jet at stoichiometry &
$Y_{CH_4,1}$ & CH$_4$ mass fraction, inner (fuel) stream\\
$S_1,S_2$ & swirl numbers, inner and outer streams &
$Y_{O_2,2}$ & O$_2$ mass fraction, outer (oxidizer) stream\\
$Re_1,Re_2$ & Reynolds numbers, inner and annular streams &
$\rho u^2A_1$, $\rho u^2A_2$ & axial fluxes of axial momentum [kg\,m/s$^2$]\\[2mm]
\multicolumn{4}{@{}l}{\textit{Model variables}}\\[1mm]
$\mathbf{x}$ & flame image, $\mathbb{R}^{512\times64\times1}$ &
$\mathbf{y}$ & standardized vector of the 15 parameters, $\mathbb{R}^{15}$\\
$\mathbf{x}_0$ & clean (un-noised) image &
$\mathbf{z}$ & latent noise vector\\
$\mathbf{x}_t$ & image at diffusion step $t$ &
$\hat{\mathbf{x}}_0$ & estimate of $\mathbf{x}_0$ reconstructed from $\mathbf{x}_t$\\
$G$ & generator &
$D$ & discriminator\\
$\phi(\mathbf{x})$ & pooled image feature vector, $\mathbb{R}^{d}$ &
$V$ & linear embedding of $\mathbf{y}$, $\mathbb{R}^{d\times15}$\\
$d$ & dimension of the pooled feature space &
$\theta$ & network weights\\
$\boldsymbol{\epsilon}$ & Gaussian noise, $\mathcal{N}(0,I)$ &
$\boldsymbol{\epsilon}_\theta$ & network prediction of $\boldsymbol{\epsilon}$\\
$\gamma(\mathbf{y}),\beta(\mathbf{y})$ & CBN scale and shift predicted from $\mathbf{y}$ &
$\mathrm{sg}(\cdot)$ & stop-gradient operator\\[2mm]
\multicolumn{4}{@{}l}{\textit{Diffusion schedule and losses}}\\[1mm]
$T$ & number of diffusion steps &
$\beta_t$ & noise variance added at step $t$\\
$\alpha_t$ & $1-\beta_t$ &
$\bar\alpha_t$ & cumulative product $\prod_{s\le t}\alpha_s$\\
$t_{\mathrm{trunc}}$ & step below which noise injection is halted &
$M(\mathbf{x}_0)$ & luminance weighting mask\\
$\theta_{\mathrm{EMA}}$ & Polyak-averaged shadow weights &
$\mathcal{L}_{\mathrm{noise}}$ & noise-prediction loss\\
$\mathcal{L}_{\mathrm{adv}}$ & adversarial loss &
$\mathcal{L}_{L1}$ & $L_1$ reconstruction loss\\
$\lambda_{L1}$ & weight of the $L_1$ anchor &
$\lambda_{x_0}$ & weight of the $\mathbf{x}_0$ reconstruction terms\\
$\alpha$ & scale factor in the dynamic weighting &
$\mu$ & EMA momentum\\
$\alpha_G,\alpha_D$ & learning rates, generator and discriminator &
$\beta_1,\beta_2$ & Adam momentum coefficients\\
$\delta$ & numerical stabilizer, $10^{-8}$ & &\\
\end{longtable}}
\vspace{-4mm}

\section{Introduction}

Heavy industries such as glass manufacturing and metallurgy increasingly rely on oxy-combustion, in which pure oxygen replaces atmospheric air as the oxidizer \cite{chen2012oxyfuel}. Eliminating the nitrogen ballast raises the adiabatic flame temperature and radiative heat transfer, increases the furnace efficiency, reduces flue gas volumes by 70--80\%, and yields a nearly pure CO$_2$/H$_2$O exhaust stream compatible with carbon capture \cite{chen2012oxyfuel,jourdaine2017}. These benefits come with extreme physical constraints: local temperatures can exceed 2500\,$^\circ$C, and macroscopic flame characteristics---such as axial length \cite{degeneve2019scaling}, radial expansion, and lift-off height \cite{degeneve2019swirl}---can directly impact process safety and furnace lifetime (burner degradation, risk of flashback or flame blow-out \cite{syred1974}), as well as process efficiency, pollutant emissions ---like nitrogen oxides (NOx)--- and product quality \cite{chen2012oxyfuel}.

Developing burners that control these phenomena requires accounting for more than ten tightly coupled operating variables (thermal power, fuel/oxidizer flow rates, momentum ratios, swirl numbers, etc.), either through expensive numerical simulations or through full-scale experimental trials of successive burner designs \cite{chen2012oxyfuel,degeneve2019scaling}.  A common diagnostic collects the OH* chemiluminescence emission from the flame, a good marker of the maximal heat release location \cite{hardalupas2004}. A single campaign consumes days of facility time and large quantities of pure oxygen, making iterative design exploration effectively impossible.

This paper addresses that bottleneck with an AI-based predictive approach: a model trained on an existing experimental dataset of 1,609 OH* 2D fields predicts the flame shape for a new set of parameters. The approach can be seen as a cross-modal continuous regression problem: learn a mapping from the tabular operating vector $\mathbf{x}\in\mathbb{R}^{15}$ to the spatial OH* emission field $\mathbf{Y}\in\mathbb{R}^{512\times64\times1}$. This enables an engineer to obtain physically accurate flame images interactively, bypassing the days-long bottlenecks of traditional methods. The central challenge is data scarcity: although 1,609 experimental samples is an uncommonly large database, it remains orders of magnitude below the regimes modern generative models were designed for, while the underlying physics exhibits sharp non-linearities such as vortex-breakdown bifurcations, where a 2\% change in swirl number can abruptly transform an attached jet flame into a lifted toroidal structure \cite{syred1974,degeneve2021liftoff}.

The contributions are twofold. First, a spatially-anchored conditional least-squares GAN (cLSGAN-L1-CBN) \cite{mirza2014} is engineered through a sequence of targeted interventions---early spatial conditioning, a projection discriminator for continuous labels \cite{miyato2018}, conditional batch normalization \cite{devries2017}, an L1 geometric anchor \cite{isola2017}, and autonomous dynamic loss balancing in the spirit of adaptive multi-task weighting---that jointly resolve the perception--distortion trade-off between geometric accuracy and texture realism. Second, a physics-guided denoising diffusion probabilistic model (DDPM) \cite{ho2020} is adapted to sparse chemiluminescence images via a luminance-weighted reconstruction objective, CoordConv coordinate channels \cite{liu2018}, Polyak-averaged weights, and a truncated-inference sampling loop. Both paradigms are validated with a common multi-metric protocol (RMSE, PSNR, SSIM, FID) on a stratified test set, and their scaling behaviour is compared on a merged training corpus.

\section{Experimental Dataset and Problem Formulation}
\subsection{Data Origin and Acquisition}
\begin{figure}[!tbp]
    \centering
    % First figure
    \begin{minipage}{0.42\textwidth}
        \includegraphics[width=\linewidth]{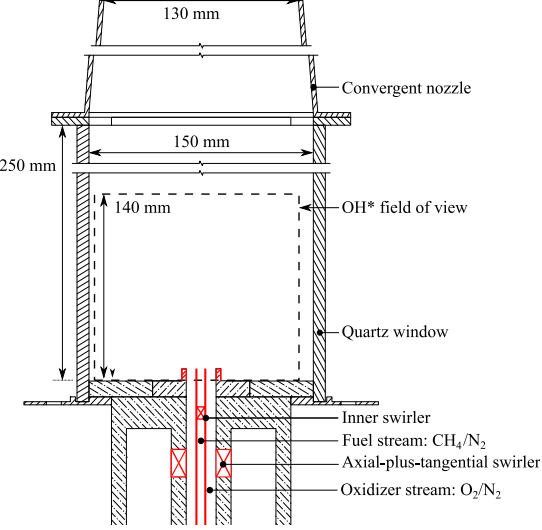}
        \caption{Schematic of the Oxytec test rig. Reproduced from \cite{degeneve2019swirl} with permission of the authors.}
        \label{fig:injector}
    \end{minipage}
    \hspace{2cm} 
    % Second figure
    \begin{minipage}{0.35\textwidth}
        \includegraphics[width=\linewidth]{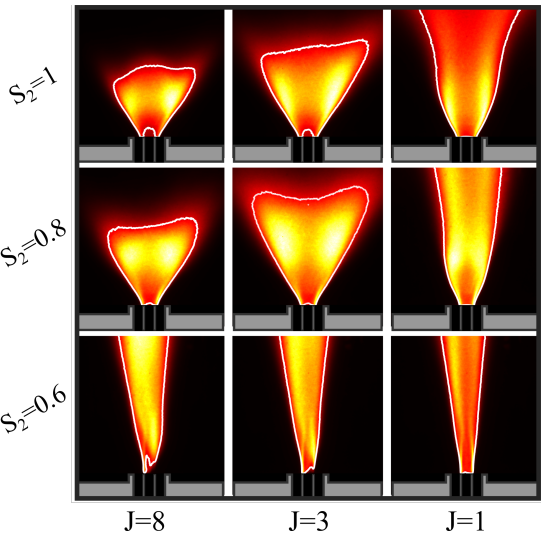}
        \caption{OH* flame images as a function of the momentum flux ratio $J$ and the outer swirl number $S_2$. The inner swirl is fixed to $S_1=0$. The Otsu contour delineating the flame front location is represented in white. Reproduced from \cite{degeneve2019swirl} with permission of the authors.}
        \label{fig:chamber}
    \end{minipage}
\end{figure}

The dataset originates from the Air Liquide, CentraleSupelec and CNRS Chair on oxy-combustion and heat transfer for energy and environment and the OXYTEC project (ANR-12-CHIN-0001) from l’Agence Nationale de la Recherche \cite{jourdaine2017,degeneve2019scaling,degeneve2019swirl}. All samples correspond to a dual-swirl non-premixed configuration produced by a coaxial injector: a central tube ($d_1=6$\,mm) carries nitrogen-diluted methane, while an annular channel ($d_2=20$\,mm) carries O$_2$-enriched air. Swirl is imposed on the inner jet by an axial swirler and on the annular jet by splitting the flow between axial and tangential inlets; a schematic of the test rig is given in Fig.~\ref{fig:injector}. Flames are stabilized in a $150\times150\times250$\,mm quartz-walled chamber, and reaction zones are imaged by OH* chemiluminescence using a 16-bit intensified CCD camera (Princeton PIMAX4) with a UV lens and a 310\,nm narrowband filter. Figure~\ref{fig:chamber} illustrates the diversity of flame topologies spanned by the database across the momentum flux ratio $J$ and the outer swirl number $S_2$, whose stabilization mechanisms are analysed in \cite{degeneve2019swirl,degeneve2021liftoff}.

Each image is paired with 15 operating and thermodynamic parameters, including the equivalent momentum ratio $J=\rho_2^{\mathrm{eq}}u_2^2/(\rho_1 u_1^2)$, global equivalence ratio $\Phi$, stoichiometric mixture and mole fractions ($Z_{st}$, $X_{st}$), inner/outer swirl numbers ($S_1$, $S_2$), Reynolds numbers ($Re_1$, $Re_2$), axial momentum fluxes, total power, quarl angle, equilibrium stoichiometric temperature, and reactant mass fractions. Otsu-filtered flame length and lift-off height are extracted per image and used exclusively to stratify data splitting.

\subsection{Preprocessing and Splitting}
Exploiting axial symmetry, the raw captures are cropped, aligned to the burner lip, and standardized to $512\times64$ grayscale half-flame tensors normalized to $[-1,1]$. The 15 tabular features are standard-scaled to zero mean and unit variance. Because extreme aerodynamic events (pronounced lift-off, widened flames) are rare but industrially critical, the two Otsu metrics are discretized into five bins each to form a stratification proxy, and the 1,609 samples are split 60/20/20 into training, validation, and test sets with balanced topological coverage. All models are implemented in TensorFlow~2.10/Keras % 
and trained on a single mobile workstation GPU (NVIDIA RTX 3500 Ada, 12\,GB VRAM).

\section{Paradigm 1: Spatially-Anchored Conditional LSGAN (cLSGAN-L1-CBN)}
\label{sec:gan}
Standard conditional GANs \cite{mirza2014} are designed for categorical labels; conditioning on continuous regression labels risks empirical-risk-minimization collapse, since virtually no training image exists at any exact coordinate of the 15-dimensional space \cite{ding2020}. Throughout this section, $\mathbf{x}\in\mathbb{R}^{512\times64\times1}$ denotes a flame image and $\mathbf{y}\in\mathbb{R}^{15}$ the standardized vector of operating parameters conditioning its generation. The architecture combines four interventions.

\textbf{Projection discriminator.} Instead of tiling conditions into the image grid on the discriminator side, the image is compressed by strided $4\times4$ convolutions and global average pooling into a feature vector $\phi(\mathbf{x})\in\mathbb{R}^{d}$, while the physical vector $\mathbf{y}$ is linearly embedded into the same space through $V\in\mathbb{R}^{d\times15}$. The conditional score is the inner product of the two, added to an unconditional realism score \cite{miyato2018}:
\begin{equation}
D(\mathbf{x},\mathbf{y})=\mathrm{Dense}_{\mathrm{uncond}}\!\big(\phi(\mathbf{x})\big)+\big\langle\,\phi(\mathbf{x}),\,V\mathbf{y}\,\big\rangle,
\end{equation}
which strongly penalizes topologies inconsistent with the requested thermodynamics.

\textbf{Least-squares objective.} With binary cross-entropy, the sigmoid-bounded discriminator saturates on this rigidly structured data and starves the generator of gradients. The LSGAN formulation \cite{mao2017} removes the sigmoid and optimizes with $\mathbf{z}$ the latent noise vector,
\begin{equation}
\mathcal{L}_G=\tfrac{1}{2}\,\mathbb{E}_{\mathbf{z},\mathbf{y}}\big[(D(G(\mathbf{z},\mathbf{y}))-1)^2\big],
\end{equation}
whose linear gradient never vanishes, restoring a stable minimax equilibrium.

\textbf{Conditional batch normalization (CBN).} To prevent the low-variance physical vector from being diluted through successive transposed convolutions, static batch-norm affine parameters are replaced at every generator resolution by condition-dependent transformations $\gamma(\mathbf{y})=W_\gamma\mathbf{y}+b_\gamma$ and $\beta(\mathbf{y})=W_\beta\mathbf{y}+b_\beta$ \cite{devries2017}, so the physics modulates every feature map.

\textbf{L1 anchor with autonomous balancing.} Adversarial losses prioritize perceptual deception over geometric correctness, so an L1 penalty $\mathcal{L}_{L1}=\mathbb{E}\|\mathbf{x}_{\mathrm{real}}-G(\mathbf{z},\mathbf{y})\|_1$ \cite{isola2017} anchors the macroscopic topology. Rather than hand-tuning a static weight (a data-leakage risk), the weight is computed each step from exponential moving averages (EMA, momentum $\mu=0.99$) of both losses, with $\alpha=1$ and $\delta=10^{-8}$,
\begin{equation}
\lambda_{L1}^{(t)}=\alpha\,\frac{\mathbb{E}[\mathcal{L}_{\mathrm{adv}}^{(t)}]}{\mathbb{E}[\mathcal{L}_{L1}^{(t)}]+\delta},\qquad
\mathcal{L}_{\mathrm{Total}}^{(t)}=\mathcal{L}_{\mathrm{LSGAN}}+\mathrm{sg}\!\big(\lambda_{L1}^{(t)}\big)\cdot\mathcal{L}_{L1},
\end{equation}
and a gradient-stop $\mathrm{sg}(\cdot)$ preventing the optimizer from gaming the coefficient. Training uses a two-time-scale update rule with dual Adam optimizers ($\beta_1=0$, $\beta_2=0.9$; $\alpha_G=8\times10^{-4}$, $\alpha_D=2\times10^{-4}$) and exponential learning-rate decay.

\section{Paradigm 2: Physics-Guided Denoising Diffusion Model}
\label{sec:ddpm}
Although the anchored GAN delivers instantaneous inference, its single-step generation struggles to reproduce the smooth scalar gradients of a fluid manifold. DDPMs \cite{ho2020} replace the single-step mapping with an iterative reverse Markov chain, a strong inductive bias for spatial continuity. The forward process uses $T=1k$ steps with a linear variance schedule ($\beta_{\mathrm{start}}=10^{-4}$, $\beta_{\mathrm{end}}=0.02$), and, writing $\alpha_t=1-\beta_t$ and $\bar\alpha_t=\prod_{s\le t}\alpha_s$, arbitrary noisy states are sampled in closed form as $\mathbf{x}_t=\sqrt{\bar\alpha_t}\,\mathbf{x}_0+\sqrt{1-\bar\alpha_t}\,\boldsymbol{\epsilon}$, where $\boldsymbol{\epsilon}\sim\mathcal{N}(0,I)$. The reverse process is parameterized by a U-Net \cite{ronneberger2015} with Swish activations; the timestep and the 15 physical features are embedded through independent 256-dimensional Swish blocks, fused into a 512-dimensional context vector, and injected at every resolution by affine channel addition. Four physics-oriented modifications are applied.

\textbf{Luminance-masked reconstruction loss.} Flame tensors are dominated by a quiescent black background, so pure noise prediction $\mathcal{L}_{\mathrm{noise}}=\|\boldsymbol{\epsilon}-\boldsymbol{\epsilon}_\theta\|_2^2$, with $\theta$ the network weights, permits unconstrained spatial drift. At each step, a clipped analytical estimate $\hat{\mathbf{x}}_0=(\mathbf{x}_t-\sqrt{1-\bar\alpha_t}\,\boldsymbol{\epsilon}_\theta)/\max(\sqrt{\bar\alpha_t},10^{-4})$ is penalized under a luminance mask $M(\mathbf{x}_0)=1+4\,\mathrm{ReLU}\!\big((\mathbf{x}_0+1)/2\big)$ that weights the flame core up to $5\times$ more than the background:
\begin{equation}
\mathcal{L}_{\mathrm{Total}}=\mathcal{L}_{\mathrm{noise}}
+50\Big(\big\langle \mathrm{sg}(M(\mathbf{x}_0))\odot\|\mathbf{x}_0-\hat{\mathbf{x}}_0\|_1\big\rangle
+\big\langle\|\mathbf{x}_0-\hat{\mathbf{x}}_0\|_1\big\rangle\Big),
\end{equation}
with $\lambda_{x_0}=50$, locking macroscopic boundaries early in the reverse process.

\textbf{CoordConv.} Standard convolutions are translation-invariant and blind to absolute position, which is problematic for constraints such as lift-off height. Normalized coordinate grids $X,Y\in[-1,1]^{512\times64}$ are concatenated to the noisy input \cite{liu2018}.

\textbf{EMA weights.} A shadow copy $\theta_{\mathrm{EMA}}$ is updated with momentum $\beta=0.999$ and used for all inference, insulating evaluation from final-batch stochasticity.

\textbf{Truncated inference.} Standard ancestral sampling injects noise down to $t\to0$, leaving granular background artifacts on sparse imagery. Setting the noise to zero below a threshold $t_{\mathrm{trunc}}$ turns the terminal segment of the chain into a deterministic refinement; a sweep over $t_{\mathrm{trunc}}\in\{10,50,100,150,200,300\}$ identifies the optimum without retraining.

\section{Results and Discussion}
\label{sec:results}
Both paradigms are developed through an ablation study: starting from a plain conditional baseline, the physical constraints introduced in Sections~\ref{sec:gan} and~\ref{sec:ddpm} are added one at a time, and each intermediate model is retrained and re-evaluated under identical conditions. This isolates the individual contribution of every intervention and identifies which mechanism resolves which failure mode, rather than reporting only the performance of the final architecture. All variants are evaluated on the stratified test set, which was held out from training in every configuration reported in this paper, including the final merged-corpus runs of Section~\ref{sec:final-models}. Evaluation uses a common metric ensemble: RMSE (pixel-wise geometric error), PSNR and SSIM (structural gradients and luminance), and FID (distributional realism), complemented by qualitative inspection. To select a representative reference case, baseline predictions were ranked by error. The highest-error 5\% tail was excluded, as those cases correspond to the pathological, sparsely-sampled operating points of Section~\ref{sec:validation}; the reference sample, Flame ID 1736, was drawn from the next 5\% band and held fixed throughout the ablation. Inspection was extended to the 50 highest-error predictions, beyond which no visible degradation appeared. GAN ablations are trained for 500 epochs and DDPM ablations for 1k epochs to isolate architectural effects.

\subsection{cLSGAN Ablation}
Table~\ref{tab:gan} summarizes the adversarial progression. The late-conditioned cGAN baseline attains a favourable FID (50.86) but collapses conditionally, largely ignoring the requested operating point (local RMSE 0.128 on the reference flame). Broadcasting the 15 features onto the input grid locks the macroscopic geometry (RMSE 0.041) at the cost of pronounced checkerboard textures (FID 89.61). Switching to the least-squares objective stabilizes the minimax gradients but cannot alone restore texture, whereas CBN without any spatial constraint produces the opposite failure: photorealistic turbulence (FID 31.97) that drifts freely across the canvas (RMSE 0.057), since its affine modulation carries no positional information. The L1 anchor is the decisive step, driving RMSE to 0.029 and PSNR to 35.58, although direct concatenation still leaves horizontal structural artifacts (FID 50.67). Combining the anchor with CBN reconciles both objectives at last---FID 31.65, RMSE 0.028, SSIM 0.965. Replacing the hand-picked weight $\lambda_{L1}=100$ with dynamic EMA balancing then yields the best perceptual score of the study (FID 30.98) at practically identical geometric accuracy; its principal value, however, is methodological, as it removes the human-tuned hyperparameter and thus any implicit test-set bias from the training loop.

\begin{table}[!tbp]
\centering
\caption{Ablation of the adversarial paradigm on the held-out test set (500 epochs per variant). Best value of each metric in bold.}
\label{tab:gan}
\small
\begin{tabular}{lcccc}
\toprule
Model & FID $\downarrow$ & RMSE $\downarrow$ & PSNR $\uparrow$ & SSIM $\uparrow$\\
\midrule
cGAN, late conditioning                       & 50.86 & 0.053 & 29.93 & 0.903\\
cGAN, early spatial conditioning              & 89.61 & 0.041 & 31.20 & 0.925\\
cLSGAN                                        & 93.89 & 0.042 & 30.26 & 0.920\\
cLSGAN + CBN                                  & 31.97 & 0.057 & 27.72 & 0.912\\
cLSGAN + L1 anchor                            & 50.67 & 0.029 & \textbf{35.58} & 0.964\\
cLSGAN + L1 + CBN (static $\lambda=100$)      & 31.65 & \textbf{0.028} & 34.70 & \textbf{0.965}\\
cLSGAN + L1 + CBN (dynamic balancing)         & \textbf{30.98} & 0.029 & 34.64 & \textbf{0.965}\\
\bottomrule
\end{tabular}
\end{table}

\subsection{DDPM Ablation and Truncated Inference}
Table~\ref{tab:ddpm} charts the diffusion progression. The vanilla conditional DDPM (Model A) fails catastrophically on sparse imagery (RMSE 0.135, FID 153.26), drifting across the canvas and hallucinating background artifacts. The luminance-masked objective (Model B) halves the RMSE to 0.065; CoordConv (Model C) eliminates blank-tensor and drifting-flame failure modes (RMSE 0.048); and EMA weights (Model D) yield the best structural configuration (RMSE 0.040, SSIM 0.903). Throughout, FID worsens (peaking at 202.47) as generative capacity is bottlenecked toward geometry---a persistent grain that motivates intervening on the sampler rather than the network. Halting stochastic injection at $t_{\mathrm{trunc}}=150$ (Table~\ref{tab:ddpm}) collapses the FID from 202.47 to 57.84 while further improving RMSE to 0.036 and SSIM to 0.936, with no retraining.

\begin{table}[!tbp]
\centering
\caption{DDPM ablation and effect of truncated inference (1k epochs per variant). Best value of each metric in bold.}
\label{tab:merged_ddpm_trunc}
\small
\begin{tabular}{lcccc}
\toprule
Model & FID $\downarrow$ & RMSE $\downarrow$ & PSNR $\uparrow$ & SSIM $\uparrow$\\
\midrule
A: Vanilla conditional DDPM                    & 153.26 & 0.135 & 19.44 & 0.561\\
B: A + luminance masking                       & 163.42 & 0.065 & 26.76 & 0.845\\
C: B + CoordConv                               & 181.53 & 0.048 & 28.40 & 0.810\\
D: C + EMA weights ($t_{\mathrm{trunc}}=0$)    & 202.47 & 0.040 & 29.44 & 0.903\\
E: D + Truncated inference ($t_{\mathrm{trunc}}=150$) & \textbf{57.84} & \textbf{0.036} & \textbf{30.97} & \textbf{0.936}\\
\bottomrule
\label{tab:ddpm}
\end{tabular}
\end{table}

\subsection{Final Models on the Merged Corpus} \label{sec:final-models}
\begin{figure}[!tbp]
\centering
\IfFileExists{FinalComparison.png}{\includegraphics[width=0.6\textwidth]{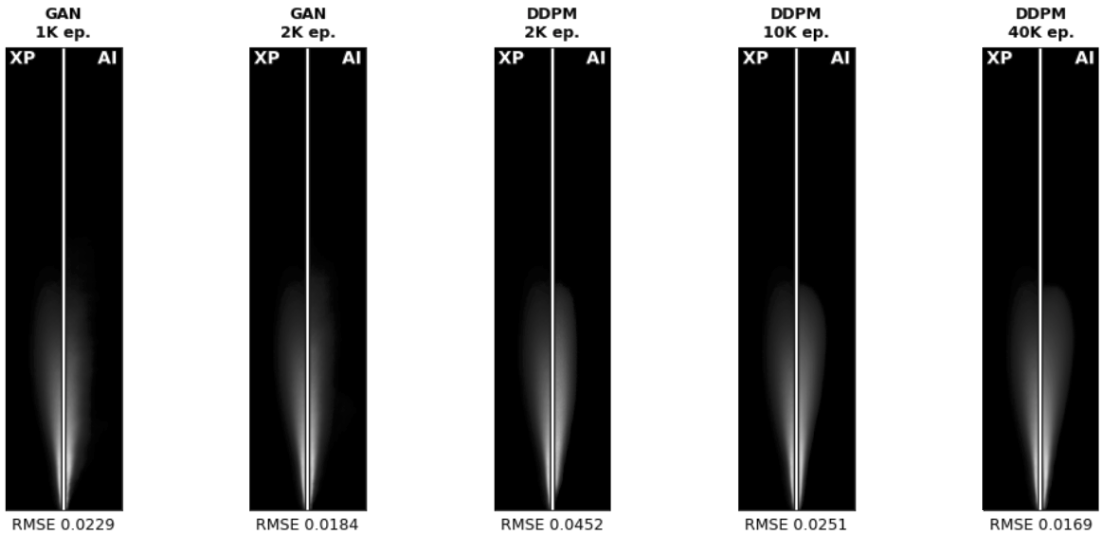}}{\fbox{\parbox[c][4cm][c]{0.8\linewidth}{\centering\small [FinalComparison.png]}}}
\caption{Finalized models on the reference flame (ID~1736). Each panel mirrors the experimental half-flame (XP, left of the seam) against the generated half-flame (AI, right), the seam being the flame axis. Columns, left to right: cLSGAN-L1-CBN at 1k and 2k epochs; DDPM at 2k, 10k, and 40k epochs. Local RMSE is annotated below each panel.}
\label{fig:final}
\end{figure}

\begin{figure}[!tbp]
\centering
\IfFileExists{mirroredTestComparison.png}{\includegraphics[width=0.7\textwidth]{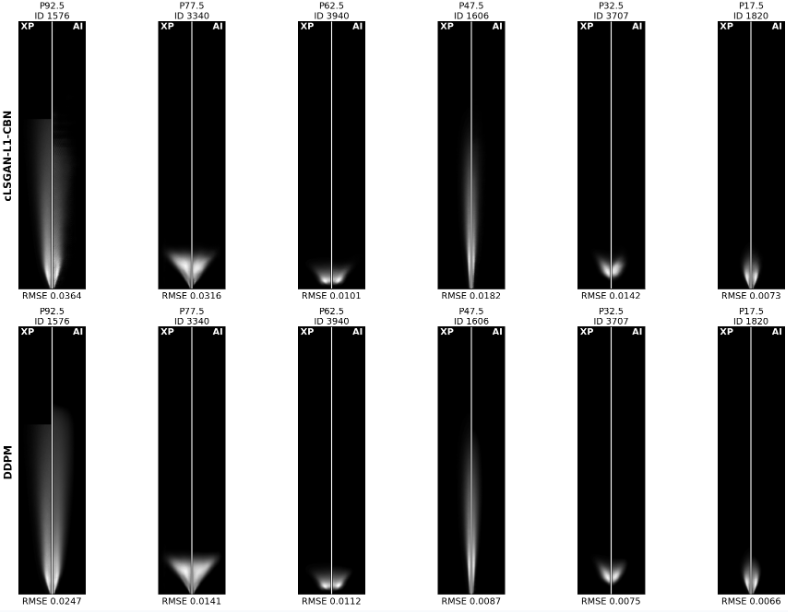}}{\fbox{\parbox[c][4cm][c]{0.8\linewidth}{\centering\small [mirroredTestComparison.png]}}}
\caption{Experimental half-image (XP) and the generated half-image (AI) across six test cases sampled from six error percentile bands (from P92.5 to P17.5). Each panel mirrors the experimental half-image against the generated half-image, the seam being the flame axis. Top row: cLSGAN-L1-CBN; bottom row: DDPM. Local RMSE is annotated below each panel.}
\label{fig:mirror}
\end{figure}

With every architectural and weighting choice locked during ablation, the training and validation sets are merged to maximize information density, and both finalized models are retrained (Table~\ref{tab:final}). The cLSGAN-L1-CBN reaches RMSE 0.025 / FID 29.68 at 1k epochs, and a further 1k epochs improve the aggregate metrics to RMSE 0.022 / FID 20.68. This numerical gain, however, does not translate into the targeted qualitative improvement: although individual predictions can sharpen locally---as on the reference flame (Fig.~\ref{fig:final}, first two columns)---the two checkpoints produce nearly identical outputs over most of the test set, and the specific limitation that motivated the extended run---internal fields that occasionally remain rigid or segmented, with textures that are not always fully realistic---persists essentially unchanged at 2k epochs. Prolonging training confirms this saturation: at 4k epochs the test-set RMSE degrades to 0.024. Extended adversarial training thus redistributes error within the low-frequency structure the metrics reward most, rather than deepening the generator's representation of the underlying flame physics; the high-frequency texture deficit is intrinsic to single-step adversarial mapping on sparse fluid data and is not resolved by training duration alone. The DDPM behaves differently: because a fresh noise tensor and random timestep are drawn at every forward pass, the network effectively never sees the same input twice and is strongly insulated against memorization. It overtakes the adversarial surrogate in geometric accuracy by 10k epochs (RMSE 0.021 versus 0.022) and reaches RMSE 0.019 / FID 20.29 / PSNR 38.33 at 40k epochs, beyond which loss tracking indicates incipient overfitting and training is halted. Qualitatively (Figs.~\ref{fig:final} and~\ref{fig:mirror}, with the corresponding operating parameters in Table~\ref{tab:testcases}), the 40K-epoch DDPM renders smooth, continuous thermodynamic gradients that the single-step generator cannot replicate (local RMSE 0.0169 on the challenging reference flame). In terms of latency, the two paradigms remain sharply differentiated: the GAN synthesizes an image in a single forward pass in about 20\,ms, three orders of magnitude faster than the full 1k-step reverse diffusion chain, which requires roughly thirty seconds per sample on the same hardware.
\begin{table}[!tbp]
\centering
\caption{Finalized surrogates trained on the merged training+validation corpus, evaluated on the test set. All DDPM rows use the truncated sampler ($t_{\mathrm{trunc}}=150$). Best value of each metric in bold.}
\label{tab:final}
\small
\begin{tabular}{lccccc}
\toprule
Final architecture & Epochs & FID $\downarrow$ & RMSE $\downarrow$ & PSNR $\uparrow$ & SSIM $\uparrow$\\
\midrule
cLSGAN-L1-CBN          & 1,000  & 29.68 & 0.025 & 35.90 & 0.971\\
cLSGAN-L1-CBN          & 2,000  & 20.68 & 0.022 & 37.96 & \textbf{0.977}\\
DDPM                   & 2,000  & 45.44 & 0.031 & 32.93 & 0.948\\
DDPM                   & 10,000 & 26.05 & 0.021 & 36.72 & 0.962\\
DDPM                  & 40,000 & \textbf{20.29} & \textbf{0.019} & \textbf{38.33} & 0.973\\
\bottomrule
\end{tabular}
\end{table}

\subsection{Persistent Outliers and Physical Sensitivity Validation} \label{sec:validation}
Across the ablations of both paradigms, a small subset of test conditions consistently remained among the poorest predictions, with errors of nearly constant magnitude regardless of architecture or training horizon, consistent with the observation of Section~\ref{sec:results} that visible degradation is confined to the top 50 highest-error cases. A nearest-neighbour analysis in the standardized 15-dimensional parameter space showed that these conditions lie in sparsely populated regions of the operational envelope, at unusually large distances from their closest training samples. Three non-mutually-exclusive factors plausibly contribute. First, coverage: with fewer than one thousand training conditions over fifteen dimensions, isolated operating points---particularly those adjacent to aerodynamic bifurcations---offer no informative neighbours from which to interpolate. Second, ground truth quality: a few reference frames carry optical acquisition artefacts, so part of the measured error penalizes imperfections of the target rather than of the prediction. Third, path dependence of the physical system: swirl-stabilized flames can exhibit hysteresis, whereby the same set of operating parameters corresponds to different stable flame topologies depending on the trajectory followed through the parameter space to reach that condition \cite{cheneau2019,candel2014,mesquita2021}. Since the acquisition history is not part of the 15-dimensional input, such conditions define a multi-valued target that no deterministic conditional mapping can resolve, and the associated error is irreducible by construction.
These persistent outliers therefore characterize the data---its sparsity and its possible path dependence---rather than a deficiency of the learned mapping, and they delimit the confidence envelope within which the surrogate should be trusted.
To verify that the models capture the governing physics rather than memorizing the corpus---a legitimate concern once training and validation sets are merged---a sensitivity study was conducted: input vectors were perturbed by small increments along individual operating variables and the generated images inspected. The synthesized topologies evolved gradually, with flame length, radial expansion, and lift-off height responding in the directions expected from swirl-stabilized combustion, rather than snapping onto the nearest training image. Both surrogates therefore interpolate the underlying dynamics of the operational envelope instead of behaving as nearest-neighbour retrieval, supporting their use for exploring unseen conditions.

\begin{table}[!tbp]
\centering
\caption{Physical and geometric parameters corresponding to flames in Fig.~\ref{fig:mirror}, ordered from highest to lowest error percentile band.}
\label{tab:testcases}
\scriptsize
% Reduce default space between columns (default is 6pt)
\setlength{\tabcolsep}{3pt} 

% Automatically scale the table to exactly fit the page width
\resizebox{\textwidth}{!}{%
\begin{tabular}{l cccccccccccccccc c}
\toprule
ID & Error band & $J$ & $S_1$ & $S_2$ & $Y_{\mathrm{CH_4},1}$ & $Y_{\mathrm{O_2},2}$ & $\Phi$ & $P$ [kW] & Quarl angle [$^\circ$] & $Z_{st}$ & $X_{st}$ & $\rho u^2 A_1$ & $\rho u^2 A_2$ & $Re_1$ & $Re_2$ & $T_{eq,st}$ [K] \\
\midrule
1576 & P92.5 & 0.623 & 0 & 0.795 & 0.953 & 0.398 & 0.897 & 32.5 & 0 & 0.0948 & 0.159 & 0.0245 & 0.122 & 12\,919 & 15\,109 & 2\,650  \\
3340 & P77.5 & 3.06 & 0.866 & 0.697 & 1 & 0.456 & 0.395 & 11 & 0 & 0.103 & 0.175 & 0.00259 & 0.0739 & 4\,228 & 11\,705 & 2\,731 \\
3940 & P62.5 & 1.49 & -0.866 & 0.953 & 1 & 0.4 & 0.646 & 15.5 & 0 & 0.0911 & 0.156 & 0.00515 & 0.0716 & 5\,963 & 11\,570 & 2\,656 \\
1606 & P47.5 & 2.96 & 0 & 0.0092 & 0.704 & 0.5 & 0.275 & 8.45 & 0 & 0.151 & 0.224 & 0.00271 & 0.0746 & 4\,132 & 11\,728 & 2\,754 \\
3707 & P32.5 & 3.99 & 0.866 & 0.701 & 1 & 0.301 & 0.528 & 9.4 & 0 & 0.0703 & 0.121 & 0.0019 & 0.0708 & 3\,625 & 11\,577 & 2\,458\\
1820 & P17.5 & 4.1 & 0 & 0.8 & 0.34 & 0.601 & 0.103 & 4.79 & 0 & 0.307 & 0.375 & 0.00308 & 0.118 & 4\,102 & 14\,615 & 2\,743\\
\bottomrule
\end{tabular}%
}
\end{table}

\section{Conclusion}
This work demonstrated that generative AI can serve as a visual surrogate for oxy-combustion flames despite a corpus of only 1,609 samples. Its central practical outcome is a redefined burner development workflow: instead of multi-day rig campaigns, an engineer adjusts the 15 operating parameters and immediately visualizes flame length, radial expansion, and lift-off. This shift shortens development time, multiplies the number of candidate designs explored within a given budget, and reserves scarce facility time and pure oxygen for the final validation of the most promising configurations.

Reproducing flame images under limited data required custom models accounting for physical constraints: for the adversarial route, early spatial conditioning, a projection discriminator, LSGAN gradients, CBN modulation, and a dynamically balanced L1 anchor resolved the perception--distortion trade-off; for the diffusion route, the same role was played by a luminance-weighted reconstruction objective, absolute coordinate awareness, EMA stabilization, and truncated deterministic sampling. The finalized DDPM achieves the best geometric accuracy and perceptual realism of the study (RMSE 0.019, FID 20.29, PSNR 38.33, SSIM 0.973) and, unlike the adversarial surrogate---whose metric gains under prolonged training leave its characteristic texture-realism deficit uncorrected and eventually reverse in geometric accuracy---benefits from long training horizons, at the cost of an iterative sampling loop of roughly thirty seconds per image on a mobile workstation GPU, three orders of magnitude slower than the adversarial surrogate, which remains the choice for strictly interactive use, generating in about 20\,ms on the same hardware. Residual errors concentrate in sparse regions of the operational envelope, with flame-topology hysteresis as a further candidate explanation, while the models respond smoothly and physically to perturbed inputs rather than memorizing the corpus. Future work could address accelerated diffusion samplers (e.g., DDIM-style schedules) to close the latency gap, a principled selection of the DDPM reconstruction weights---the luminance-mask amplitude and the fixed coefficient of Eq.~(4), which could be balanced dynamically as for the adversarial L1 anchor---uncertainty quantification near aerodynamic bifurcations, targeted acquisition in the identified sparse regions, and extension to time-resolved flame dynamics.

\section*{Acknowledgements}
 The authors thank the Air Liquide, CentraleSupelec and CNRS Chair on oxy-combustion and heat transfer for energy and environment and the OXYTEC project (ANR-12-CHIN-0001) from l’Agence Nationale de la Recherche for providing the experimental database.


\begin{thebibliography}{99}\small\setlength{\itemsep}{0pt}\setlength{\parskip}{0pt}

\bibitem{chen2012oxyfuel} L. Chen, S. Z. Yong, and A. F. Ghoniem, ``Oxy-fuel combustion of pulverized coal: Characterization, fundamentals, stabilization and CFD modeling,'' \textit{Progress in Energy and Combustion Science}, vol. 38, no. 2, pp. 156--214, 2012.

\bibitem{syred1974} N. Syred and J. M. Be\'er, ``Combustion in swirling flows: A review,'' \textit{Combustion and Flame}, vol. 23, no. 2, pp. 143--201, 1974.

\bibitem{hardalupas2004} Y. Hardalupas and M. Orain, ``Local measurements of the time-dependent heat release rate and equivalence ratio using chemiluminescent emission from a flame,'' \textit{Combustion and Flame}, vol. 139, no. 3, pp. 188--207, 2004.



\bibitem{jourdaine2017} P. Jourdaine, C. Mirat, J. Caudal, A. Lo, and T. Schuller, ``A comparison between the stabilization of premixed swirling CO$_2$-diluted methane oxy-flames and methane/air flames,'' \textit{Fuel}, vol. 201, pp. 156--164, 2017.

\bibitem{degeneve2019scaling} A. Degen\`eve, R. Vicquelin, C. Mirat, B. Labegorre, P. Jourdaine, J. Caudal, and T. Schuller, ``Scaling relations for the length of coaxial oxy-flames with and without swirl,'' \textit{Proceedings of the Combustion Institute}, vol. 37, no. 4, pp. 4563--4570, 2019.

\bibitem{degeneve2019swirl} A. Degen\`eve, C. Mirat, J. Caudal, R. Vicquelin, and T. Schuller, ``Effects of swirl on the stabilization of non-premixed oxygen-enriched flames above coaxial injectors,'' \textit{Journal of Engineering for Gas Turbines and Power}, vol. 141, no. 12, 121018, 2019.

\bibitem{degeneve2021liftoff} A. Degen\`eve, R. Vicquelin, C. Mirat, J. Caudal, T. Schuller, ``Impact of co- and counter-swirl on flow recirculation and liftoff of non-premixed oxy-flames above coaxial injectors,'' \textit{Proceedings of the Combustion Institute}, vol. 38, no. 4, pp. 5501--5508, 2021.


% 



\bibitem{mirza2014} M. Mirza and S. Osindero, ``Conditional generative adversarial nets,'' arXiv preprint arXiv:1411.1784, 2014.

\bibitem{ding2020} X. Ding, Y. Wang, Z. Xu, W. J. Welch, and Z. J. Wang, ``CcGAN: Continuous conditional generative adversarial networks for image generation,'' in \textit{Int. Conf. on Learning Representations (ICLR)}, 2021.

\bibitem{devries2017} H. De Vries, F. Strub, J. Mary, H. Larochelle, O. Pietquin, and A. C. Courville, ``Modulating early visual processing by language,'' in \textit{Advances in Neural Information Processing Systems}, vol. 30, 2017.

\bibitem{miyato2018} T. Miyato and M. Koyama, ``cGANs with projection discriminator,'' in \textit{Int. Conf. on Learning Representations (ICLR)}, 2018.

\bibitem{mao2017} X. Mao, Q. Li, H. Xie, R. Y. Lau, Z. Wang, and S. Paul Smolley, ``Least squares generative adversarial networks,'' in \textit{Proc. IEEE Int. Conf. on Computer Vision}, 2017, pp. 2794--2802.

\bibitem{isola2017} P. Isola, J.-Y. Zhu, T. Zhou, and A. A. Efros, ``Image-to-image translation with conditional adversarial networks,'' in \textit{Proc. IEEE Conf. on Computer Vision and Pattern Recognition}, 2017, pp. 1125--1134.







\bibitem{liu2018} R. Liu, J. Lehman, P. Molino, F. Petroski Such, E. Frank, A. Sergeev, and J. Yosinski, ``An intriguing failing of convolutional neural networks and the CoordConv solution,'' in \textit{Advances in Neural Information Processing Systems}, vol. 31, 2018.



\bibitem{ho2020} J. Ho, A. Jain, and P. Abbeel, ``Denoising diffusion probabilistic models,'' in \textit{Advances in Neural Information Processing Systems}, vol. 33, 2020, pp. 6840--6851.

\bibitem{ronneberger2015} O. Ronneberger, P. Fischer, and T. Brox, ``U-Net: Convolutional networks for biomedical image segmentation,'' in \textit{Medical Image Computing and Computer-Assisted Intervention (MICCAI)}, Springer, 2015, pp. 234--241.





\bibitem{cheneau2019} B. Cheneau, A. Vié, and S. Ducruix, ``Characterization of the hysteresis cycle in a two-stage liquid-fueled swirled burner through numerical simulation,'' \textit{Proceedings of the Combustion Institute}, vol. 37, pp. 5245--5253, 2019.

\bibitem{candel2014} S. Candel, D. Durox, T. Schuller, J.-F. Bourgouin, and J. P. Moeck, ``Dynamics of swirling flames,'' \textit{Annual Review of Fluid Mechanics}, vol. 46, pp. 147--173, 2014.

\bibitem{mesquita2021} L. C. C. Mesquita, A. Vi\'e, L. Zimmer, and S. Ducruix, ``Numerical analysis of flame shape bifurcation in a two-stage swirled liquid burner using Large Eddy Simulation,'' \textit{Proceedings of the Combustion Institute}, vol. 38, no. 4, pp. 5971--5978, 2021.

\end{thebibliography}
\end{document}